\documentclass[11pt]{article}

\usepackage[preprint]{acl}

\usepackage{times}
\usepackage{latexsym}

\usepackage[T1]{fontenc}

\usepackage[utf8]{inputenc}

\usepackage{microtype}

\usepackage{inconsolata}

\usepackage{graphicx}

\usepackage{enumitem}
\usepackage{booktabs}
\usepackage{subcaption}
\usepackage{wrapfig}
\usepackage{multirow}
\usepackage{makecell}
\usepackage[table]{xcolor}
\usepackage{multicol}
\usepackage{float}
\usepackage{placeins}
\usepackage[ruled,vlined,linesnumbered]{algorithm2e}
\SetKwInput{KwInit}{Init}
\SetKwInput{KwParams}{Hyperparams}
\definecolor{paleaqua}{rgb}{0.74, 0.83, 0.9}
\usepackage{amsmath} 

\title{GMTRouter: Personalized LLM Router over Multi-turn User Interactions}

\author{
Yihang Sun\thanks{Equal contribution.} \and
Encheng Xie\footnotemark[1] \and
Tao Feng \and
Jiaxuan You \\
University of Illinois Urbana-Champaign \\
\texttt{\{yihangs, encheng2, taofeng2, jiaxuan\}@illinois.edu} \\
\small\textit{Accepted to Findings of the Association for Computational Linguistics: EMNLP 2026.}
}

\begin{document}
\maketitle
\begin{abstract}
Large Language Model (LLM) routing has demonstrated strong capability in balancing response quality with computational cost. As users exhibit diverse preferences, personalization has attracted increasing attention in LLM routing, since even identical queries may require different models to generate responses tailored to individual needs. However, existing approaches are not fully personalized and often fail to faithfully capture the complex interactions between users and LLMs. Moreover, user preference data is typically scarce and inconsistent in format, which limits the effectiveness of methods that directly leverage user-specific data. To address these challenges, we propose \textit{GMTRouter}, which represents multi-turn user–LLM interactions as a heterogeneous graph with five node types: user, LLM, query, response and turn, thereby maximally preserving the rich relational structure of the interaction. Through a lightweight inductive graph learning framework combined with a tailored user-conditioned graph sampling mechanism, \textit{GMTRouter} learns to capture user preferences from few-shot data, enabling effective personalization. Extensive experiments demonstrate that \textit{GMTRouter} outperforms the strongest baselines, achieving up to a 0.108 absolute improvement in accuracy and a 0.124 improvement in AUC. More importantly, we further demonstrate that \textit{GMTRouter} can adapt to new users using only few-shot data, without extensive fine-tuning.
\end{abstract}

\section{Introduction}
With the rapid development of the field of Large Language Models (LLMs), an increasing number of models with varying sizes, computational costs, and domain expertise have become available \citep{singhal2023large,luo2022biogpt}. This makes LLM routing particularly important, as it enables the recommendation of appropriate LLMs for diverse user queries while balancing response quality with computational cost \citep{vsakota2024fly,stripelis2024polyrouter}. Such routing techniques are increasingly adopted in modern LLMs, including GPT-5 \citep{openai2025gpt5system}. At the same time, as more users engage with LLM routing services, differences in individual preferences become increasingly prominent: even identical queries may require different models to generate responses tailored to each user \citep{Li2024Personalized,Salehi2024ViPer:}. Therefore, this paper aims to highlight a pressing research question: \textit{Can we design a personalized routing framework that aligns LLM selection with individual user preferences based on their interaction histories?}

Existing research has proposed various architectures for LLM routing frameworks: FrugalGPT introduces a BERT-based router that determines whether to switch to a larger LLM \citep{frugalgpt}, while C2MAB-V constructs a bandit-based router to balance exploration and exploitation when selecting an LLM \citep{dai2024cost}. GraphRouter formulates routing as a node classification task over a graph of queries, tasks, and LLMs \citep{feng2024graphrouter}. However, existing methods largely overlook the importance of extracting structured preference information from users’ interaction histories: they are not fully personalized and often fail to faithfully model multi-turn conversations between users and LLMs, which represent the most common form of user–LLM interaction in real-world scenarios \citep{Zhang2025A,Li2025Beyond}. Moreover, in real-world scenarios, the preference data provided by a single user is typically scarce and inconsistent in format \citep{Escamocher2024Interactive,Li2024Debiased}. This makes it challenging for methods that directly leverage user-specific data to learn user profiles \citep{salemi2024optimization,Gao2024Aligning} or use such data as a retrieval source to support routing \citep{au2025personalized}, thereby limiting their effectiveness.

\begin{figure*}[t]
\begin{center}
\includegraphics[width=\linewidth]{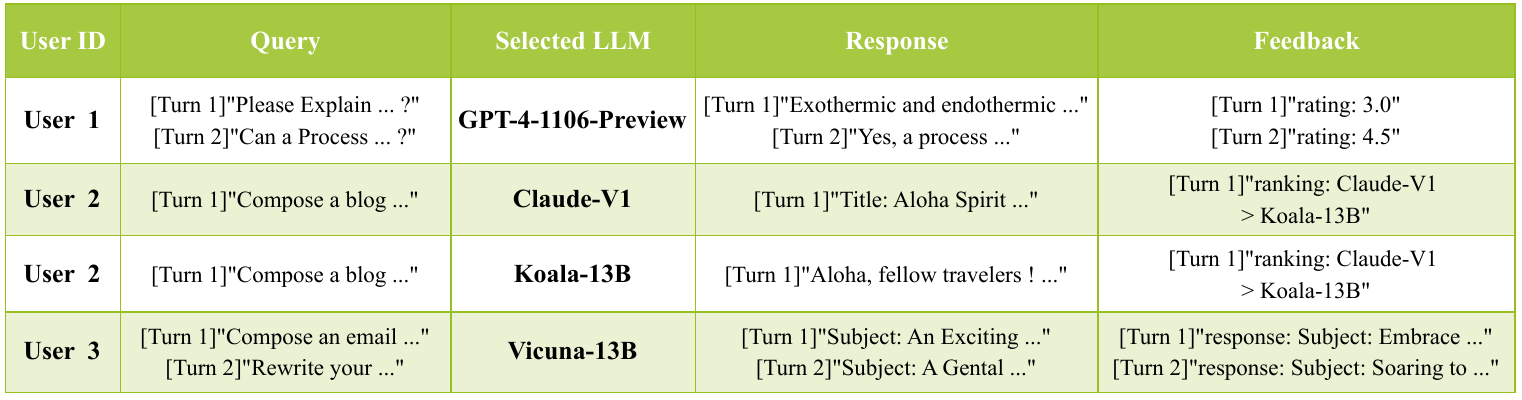}
\end{center}
\vspace{-10pt}
\caption{\textbf{Multi-turn user-LLM Interaction History Table.} Each row captures a multi-turn interaction with associated user feedback. User feedback can take various forms, including ratings, rankings, ground-truth responses.}
\label{fig:task define}
\vspace{-12pt}
\end{figure*}

To address these challenges, we introduce \textbf{GMTRouter} (\textbf{G}raph \textbf{M}ulti-\textbf{T}urn \textbf{Router}), a heterogeneous graph-based LLM router based on multi-turn user interactions for personalized LLM routing. GMTRouter first explicitly identifies the key entities within the user–LLM interaction process: users, LLMs, queries, and responses. By modeling these entities as different types of nodes and encoding their textual information into node embeddings, it maximally preserves the semantic information from the original data. To faithfully model the relational structure of multi-turn user–LLM interactions, GMTRouter organizes these diverse node types into a heterogeneous graph that captures complex relational dependencies. Each single-turn interaction is treated as a fundamental unit, and a virtual node, referred to as a \textit{turn node}, is introduced to aggregate local information within each interaction round. We further transform user preference feedback into node features, enabling preference information to propagate across the graph. Moreover, rather than training the model directly on large historical datasets, GMTRouter employs a novel \textbf{user-conditioned graph sampling} mechanism in conjunction with an inductive graph learning framework to \textbf{enhance the model’s ability to capture user preferences from few-shot data}. This design allows effective test-time personalization even under sparse interaction histories, such as cold-start scenarios involving new users.
In summary, our main contributions are as follows:

\begin{itemize}[leftmargin=*, nosep]
\item To our knowledge, we are among the first to introduce a personalized LLM routing task based on multi-turn user interactions, providing new insights for this rapidly growing research field.
\item We propose a novel personalized LLM routing framework, which faithfully models user–LLM interactions as a heterogeneous graph, and learns to capture user preferences from few-shot data within a lightweight inductive framework.
\item Through experiments on multiple datasets spanning diverse tasks, GMTRouter outperforms the strongest baselines, achieving up to a 0.108 absolute improvement in accuracy and a 0.124 improvement in AUC. Moreover, we demonstrate that our method can efficiently adapt to unseen users with only a few interaction examples, without requiring retraining.
\end{itemize}

\section{Preliminaries}

\subsection{Task Formulation}
We introduce the personalized LLM routing task in this section.
We focus on the multi-turn interaction scenario between users and LLMs with feedback \citep{Wang2023MINT:,Shi2024WildFeedback:}. Within a dialogue session, a user $u$ repeatedly interacts with an LLM $m$: in each turn, the user issues a query $q$, the LLM provides a response $r$, and the user in turn supplies a piece of feedback $f$. Such feedback can take multiple forms, including: (1) scalar scores (e.g., numerical ratings), \citep{wang2023helpsteer,Wang2024HelpSteer2:}; (2) preference rankings (e.g., choosing among multiple responses), \citep{Yang2024Aligning,sun2025premium}; (3) ground-truth responses (e.g., directly providing the correct answer) \citep{Gao2024Aligning,salemi2024optimization}.
We structure these interactions into an \textbf{Interaction History Table}, illustrated in Figure~\ref{fig:task define}, where each entry records the user ID, the selected LLM, the multi-turn queries and generated responses, and the corresponding user feedback, thereby maximally preserving the rich relational information of the interaction.

Our personalized LLM routing task is then modeled as follows:
Given $m$ users $\{u_1, \ldots, u_m\}$ and $n$ LLM candidates $\{m_1, \ldots, m_n\}$, as well as their historical interaction records:\[
\mathcal{H} = \{(u_i, m_i, \{(q^{(t)}, r^{(t)}, f^{(t)})\}_{t=1}^{T_i})\},
\]
where $u_i$ is the user, $m_i$ is the selected LLM, and each record contains a multi-turn sequence of queries $q^{(t)}$, responses $r^{(t)}$, and feedback $f^{(t)}$ for $t=1, \ldots, T_i$. 
When a user $u$ raises a new query $q$, the router is required to select an LLM $m \in \{m_1, \ldots, m_n\}$ to generate a response $r$ that best aligns with the user preferences, which is measured through the feedback $f$ provided by the user.

\begin{table}[t]
    \centering
    \small
    \caption{\textbf{The consistency of LLM preferences between users is substantially lower than the within-user consistency.} The self-consistency score is substantially higher than the spearman scores computed across different users.}
    \vspace{-10pt}
    \begin{tabular}{c c c c c}
        \toprule
        Metric & Self & Global & Intra-cluster & Inter-cluster \\
        \midrule
        Spearman & 0.793 & 0.524 & 0.573 & 0.442 \\
        Percent & 100\% & 66.0\% & 72.3\% & 55.7\% \\
        \bottomrule
    \end{tabular}
    \label{tab:spearman-summary}
\end{table}
\begin{figure}
    \centering
    \includegraphics[width=\linewidth]{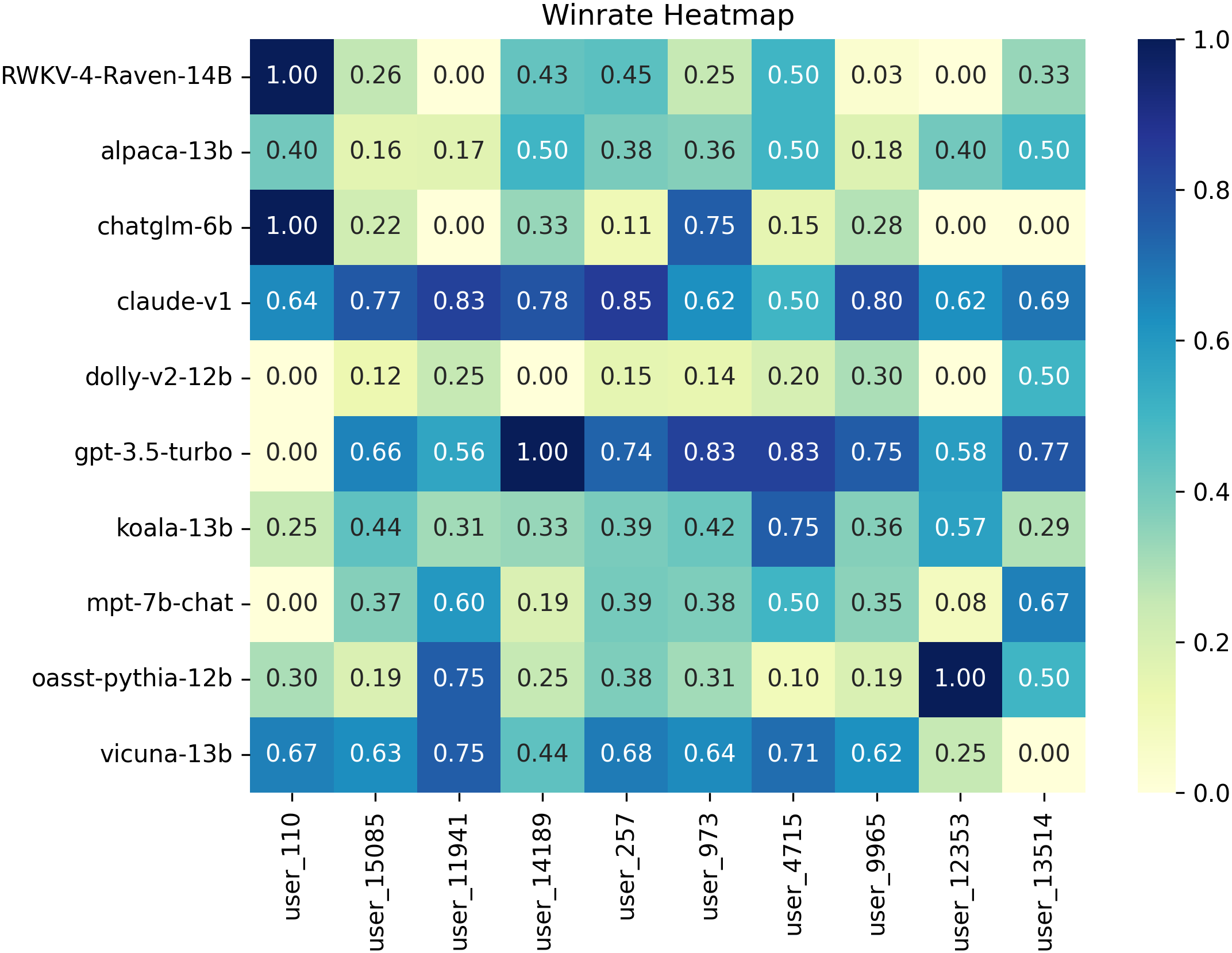}
    \vspace{-10pt}
    \caption{\textbf{Significant differences exist in LLM preferences across users.} The figure shows a heatmap of win rates for the 10 most popular LLMs across 10 active users in ChatBot Arena. Variation in win-rate patterns across rows illustrates that different users rank the same LLMs differently.}
    \label{fig:heatmap}
    \vspace{-15pt}
\end{figure}

\begin{figure*}[t]
\begin{center}
\includegraphics[width=\textwidth]{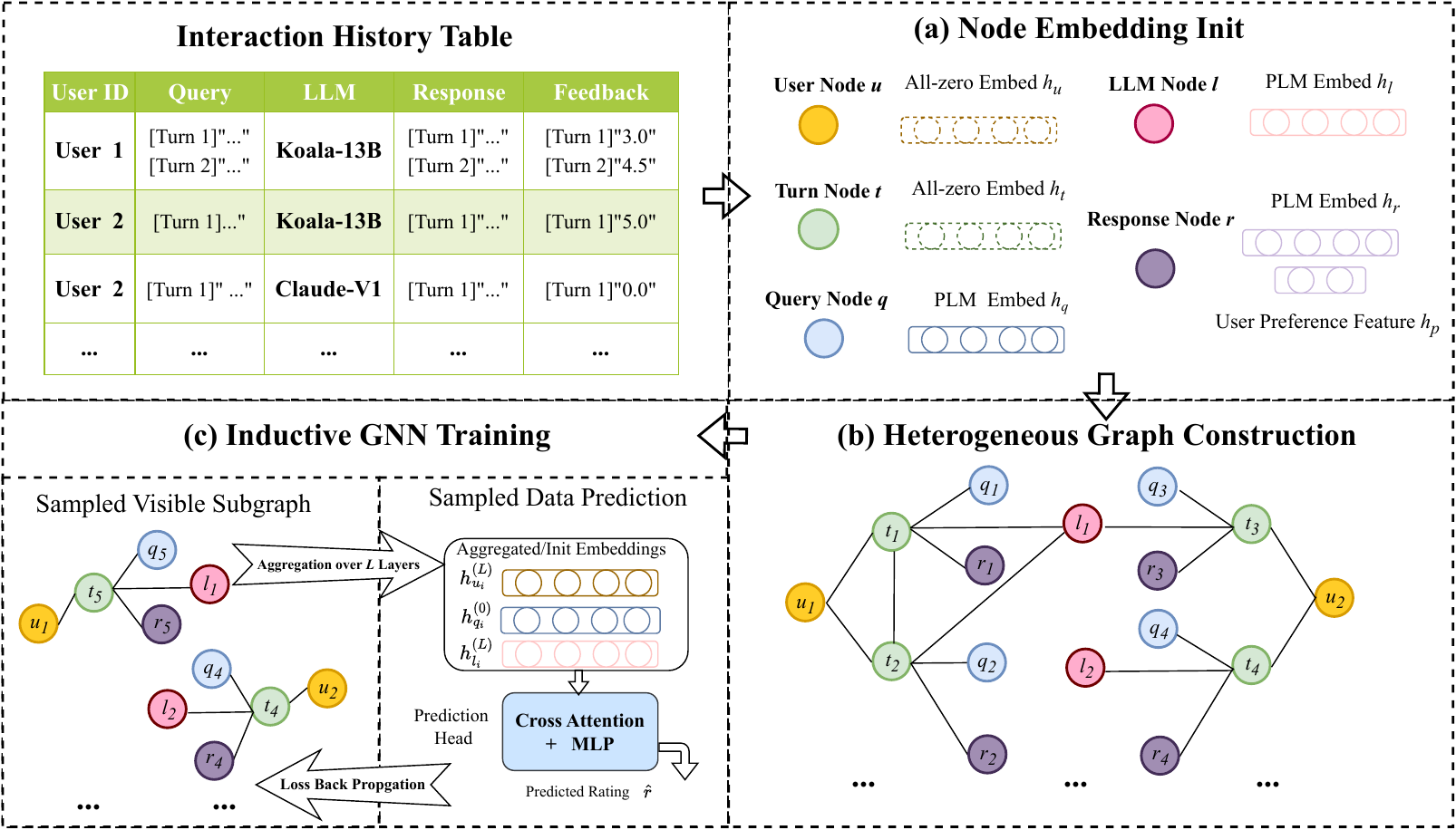}
\end{center}
\vspace{-10pt}
\caption{\textbf{Overview of GMTRouter.} (a) GMTRouter extracts users, LLMs, queries, and responses from the Interaction History Table. Query, response, and LLM nodes are initialized with PLM embeddings, while user feedback is encoded as a preference feature on response nodes. (b) It further introduces turn nodes and organizes these entities into a heterogeneous graph to model the relational structure of user--LLM interactions. (c) Within a lightweight inductive framework, GMTRouter learns to infer user preferences from few-shot interaction histories.}
\label{fig:pipeline}
\vspace{-12pt}
\end{figure*}

\subsection{Motivation}\label{sec:2.2}

In this section, we highlight the significant differences in LLM preferences across users in the real world \citep{Chevi2025How,Wang2024Arithmetic}, emphasizing the importance of personalized LLM routing for enhancing user experience. We use the ChatBot Arena dataset \citep{Chiang2024Chatbot} to illustrate our findings, which contains extensive anonymized multi-turn conversations from numerous users, with pairwise human preference labels between various LLMs, enabling the study of real-world user–LLM interactions. From this dataset, we select 10 active users, each with at least 50 records and 1,480 records in total, for detailed analysis. For each user, we randomly split their data into two halves and compute the win rates of each LLM within each half. We use Spearman correlation to quantify the consistency of preference rankings over LLMs \citep{de2016comparing,hauke2011comparison}. We then compute the Spearman correlation between the two halves to quantify their self-consistency in preferences over LLMs \citep{Chevi2025How,Jiang2025Know}, reporting the average as a baseline for comparison with inter-user preference consistency. Next, based on the similarity of queries in each user’s interaction history, we cluster users into three groups \citep{Zeng2024Personalized,Li2025LLM-Guided}, and compute pairwise Spearman correlation scores among users globally, within clusters, and across clusters \citep{Cavallo2019Functional,de2016comparing}, reporting the corresponding averages as summarized in Table \ref{tab:spearman-summary}. We observe that global consistency in LLM preferences among users is substantially lower than individual self-consistency, reaching only 65.99\% of the latter. Even within the same cluster, the Spearman score is only 72.28\% of the self-consistency, highlighting the diversity of user preferences toward LLMs \citep{sun2025premium, salemi2024optimization}. To further visualize these differences, we select the 10 most frequently used models across these 10 users and present a win-rate heatmap in Figure \ref{fig:heatmap}, offering an intuitive depiction of the variability in user preferences. 
Therefore, to address the substantial inconsistency of LLM preferences across users, we propose \textbf{GMTRouter}, a framework that enables the personalized recommendation of suitable LLMs tailored to each user’s individual preferences.
\section{GMTRouter: Router Over Multi-turn User Interactions}
\paragraph{Method Overview} As shown in Figure \ref{fig:pipeline}, \textsc{GMTRouter} operates in three stages:
(a) \textit{Node Embedding Initialization}: it extracts users, LLMs, queries, and responses from the Interaction History Table. Query, response, and LLM nodes are initialized with PLM embeddings, while user feedback is encoded as a preference feature on response nodes.
(b) \textit{Heterogeneous Graph Construction}: it further introduces turn nodes and organizes the five node types into a heterogeneous graph that captures the relational structure of multi-turn user--LLM interactions.
(c) \textit{Inductive GNN Training}: Finally, we employ a user-conditioned graph sampling mechanism and an inductive graph training framework to learn user preferences from few-shot data, enabling effective personalization under sparse interaction histories.

\subsection{Node Embedding Initialization.}
First, our framework focuses on comprehensively extracting the information of various entities involved in the user–LLM interaction process from the Interaction History Table, along with their relational structures. 
As illustrated in part (a) of Figure \ref{fig:pipeline}, we first extract four types of observed entities: user \(u\), LLM \(m\), query \(q\), and response \(r\), and formalize them as four corresponding node types. For LLM, query, and response nodes, we use a pretrained language model (PLM) to obtain their initial node embeddings \citep{wang2022text,wang2023pre}, thereby preserving the semantic information from the original data. Specifically, we encode the query and response texts as their initial embeddings, denoted as $h_q$ and $h_r$. In addition, we transform various forms of user feedback into numerical ratings and project them into a User Preference Feature $h_p$, which serves as another attribute on the response nodes. Concretely, ranking feedback is discretized into numerical ratings to ensure that higher-ranked responses receive higher scores \citep{Banditwattanawong2025Unbiased}; for ground-truth response feedback, we compute the geometric distance between the embeddings of the ground-truth and the generated response as the rating criterion \citep{salemi2024optimization}. For LLM nodes, instead of simply using their names or IDs \citep{ding2024hybrid,frugalgpt}, we encode the model overviews provided by AI/ML API platforms as their node embeddings $h_m$, which typically include key information such as model size, usage cost, and domain-specific capabilities, thereby enriching the node embeddings with important background knowledge. Finally, for user nodes, we do not assume the existence of text-based user profiles, as such information is often scarce and noisy in real-world applications \citep{Su2024Cross-domain,Alzubaidi2023A}; therefore, we initialize user embeddings $h_u$ as zero vectors.

\subsection{Heterogeneous Graph Construction.}
Next, we organize these nodes into a heterogeneous graph to model the relational structure of user–LLM interactions \citep{Zhang2025Anomaly,Schlichtkrull2017Modeling}. We consider each single-round user–LLM interaction as a fundamental unit and introduce a kind of virtual node, \textit{the turn node}, to aggregate the information within each interaction round. As illustrated in part (b) of Figure \ref{fig:pipeline}, within each interaction round, the associated user, LLM, query, and response nodes are connected to a corresponding turn node that aggregates information from that round. For multi-turn conversations, the turn nodes corresponding to each round are sequentially connected in dialogue order, facilitating information propagation across turns. The turn node embedding $h_t$ is initialized as a zero vector. The resulting heterogeneous graph captures the rich relational dependencies inherent in user–LLM interactions, where turn nodes aggregate local information within each dialogue round and propagate it to user nodes, thereby facilitating the global aggregation of user preference information.

\subsection{Inductive GNN Training}

After constructing the user–LLM interaction histories into a heterogeneous graph, we train our GNN model on it. Notably, GMTRouter is a \textbf{general framework} that can incorporate any heterogeneous GNN as its backbone. We denote the GNN backbone used in our method as $\textit{GNN}$ throughout the rest of the paper.
To address scenarios with sparse user history \citep{Su2024Cross-domain}, instead of training the model directly on large amounts of historical data \citep{Lin2021Deep-profiling:,Wang2025User}, we employ an inductive framework along with \textbf{user-conditioned graph sampling} during training, enabling GMTRouter to \textbf{capture a user’s preferences from only a few interaction records.}

\paragraph{User-conditioned Graph Sampling}
 As illustrated in the left of (c) in Figure \ref{fig:pipeline}, during each training epoch, we sample $k$ interaction histories, where each history corresponds to one user--query instance with candidate LLM responses and feedback, for each user to construct a visible subgraph $\mathcal{G}_\text{sub}$ from the heterogeneous graph for message passing, and further sample data outside the visible subgraph as the prediction targets. We then use only these small sampled visible subgraphs and perform message aggregation separately for each user to update the node embeddings. Formally, at layer $l$, node $v$'s embedding $h_v^{(l)}$ is updated by aggregating messages from its neighbors via the backbone GNN’s message-passing mechanism:
\begin{equation}
h_v^{(l)} = \text{Norm} \left( \text{Dropout} \left( \text{\textit{GNN}}^{(l)}(h^{(l-1)}_v, \mathcal{G}_\text{sub} ) \right) \right)
\end{equation}

This restriction on the amount of data involved in message passing encourages the model to learn how to infer user preferences from very limited signals and to generalize efficiently to new users.

\paragraph{LLM Routing with a Prediction Head.} After completing $L$ layers of message aggregation, we obtain the updated node representations $h^{(L)}$. We then employ a \textbf{Prediction Head} module $f_{\text{pred}}$ for preference prediction. As illustrated in the right of (c) in Figure~\ref{fig:pipeline}, the Prediction Head takes the updated user embedding $h^{(L)}_u$, LLM embedding $h^{(L)}_m$, and the query embedding $h^{(0)}q$ from PLM as input. It applies a cross-attention module, where the LLM embedding attends to the fused user–query context to extract relevant preference signals. The module outputs a scalar score $s_{u,q,m}$ for each LLM candidate, representing the likelihood that user $u$ would prefer $m$ to answer query:
\begin{equation} s_{u,q,m} = f_{\text{pred}}(h_u^{(L)}, h_q^{(0)}, h_m^{(L)}) \end{equation}

These scores are then used to rank LLM candidates under the same $(u, q)$ condition. We normalize both the predicted scores and the ground-truth ratings, and apply a criterion function to compute the training loss, which is subsequently used to update the model parameters.

During inference, when a user raises a new query, we first sample $k$ interaction histories of that user to construct the visible subgraph and update the node embeddings. Then, the LLM candidate is selected from the candidate set $\mathcal{M}$ as the one with the highest predicted score:
\begin{equation}
m^* = \arg\max_{m \in \mathcal{M}} f_{\text{pred}}(h_u^{(L)}, h_q^{(0)}, h_m^{(L)})
\end{equation}

\section{Experiment Setup}
\subsection{Datasets and data processing} 

We evaluate our approach on five datasets covering diverse tasks, spanning both real-world and synthetic settings. The two real-world datasets are: \textbf{(1) HUMAINE \citep{petrova2026unpackinghumanpreferencellms}}, which provides controlled, demographically stratified multi-turn interactions with direct human preference judgments, and \textbf{(2) ChatBot Arena \citep{Chiang2024Chatbot}}, which captures large-scale, in-the-wild multi-turn interactions and crowdsourced user preferences. The remaining three are synthetic datasets: \textbf{(3) MT-Bench \citep{zheng2023judging}}, assessing multi-turn reasoning and conversational ability; \textbf{(4) GSM8K \citep{cobbe2021gsm8k}}, focused on grade school math problem solving; and \textbf{(5) MMLU \citep{hendrycks2021ethics,hendryckstest2021}}, measuring general knowledge and multi-domain reasoning. Detailed descriptions of the datasets and preprocessing are provided in Appendix \ref{Sec: Dataset Details}.


\paragraph{Data Processing}
For HUMAINE and ChatBot Arena, we convert the pairwise human preference outcomes into ratings for the corresponding responses. For HUMAINE, persistent participant IDs are unavailable, so following its focus on demographic preference heterogeneity, we group records with the same demographic profile into a preference cohort and treat each cohort as one ``user''.

For the three synthetic datasets, we adopt the data collected in \citealt{routellm2024}, which generated responses to all questions using ``GPT-4-1106-preview'' \citep{achiam2023gpt} and ``Mixtral-8x7B-Instruct-v0.1'' \citep{Jiang2024Mixtral}. Based on this, we convert these datasets into multi-user personalized datasets. Specifically, for each response, we consider the following four dimensions:
(a) Quality: For open-ended questions, we use the GPT-4 scores provided by \citealp{routellm2024}; for objective questions, we directly evaluate the correctness.
(b) Cost: We calculate the cost of generating each response based on the API pricing provided by the AI/ML API platform.
(c) Response Length: We compute the token length of each response using the Contriever tokenizer \citep{izacard2021contriever}.
(d) Rare Words: We count the number of rare words in each response using the \textit{wordfreq} package \citep{robyn_speer_2022_7199437}.
We obtain the final rating of a response by computing a weighted sum of these four metrics. Different users are assigned different weightings to simulate diverse preferences over response quality, cost sensitivity, verbosity, and language complexity \citep{feng2024graphrouter,feng2025fusing}. The specific weights used are provided in Appendix \ref{sec:user weight}.

\paragraph{Data Splitting} For all datasets, we partition the data into training, validation, and test sets with a 7:1:2 ratio. For the new-user evaluation, we further adopt an additional splitting strategy: we sample 30\% of the users and restrict their data to the validation and test sets only, in order to evaluate the generalization ability of our method to new users unseen during training.

\begin{table*}[t]
\centering
\caption{\textbf{GMTRouter consistently outperforms baselines across all datasets.} 
Bold and underline denote the best and second-best results. The results are averaged over multiple runs.}
\label{tab:overview}
\vspace{-10pt}
\resizebox{\textwidth}{!}{%
\begin{tabular}{c||cc|cc|cc|cc|cc}
\Xhline{1pt}
\rowcolor{paleaqua}
\textbf{Method} & \multicolumn{2}{c|}{\textbf{HUMAINE}} & \multicolumn{2}{c|}{\textbf{ChatBot Arena}} & \multicolumn{2}{c|}{\textbf{MT-Bench}} & \multicolumn{2}{c|}{\textbf{GSM8K}} & \multicolumn{2}{c}{\textbf{MMLU}}\\
\rowcolor{paleaqua}
 & ACC & AUC & ACC & AUC & ACC & AUC & ACC & AUC & ACC & AUC \\
\hline\hline
\textbf{Vanilla LLM} & 0.526 & 0.544 & 0.525 & 0.741 & 0.481 & 0.457 & 0.546 & 0.533 & 0.473 & 0.475
\\
\textbf{Personalized LLM} & 0.514 & 0.548 & 0.646 & 0.780 & 0.437 & 0.491 & 0.553 & 0.536 & 0.675 & 0.678 
\\
\textbf{GraphRouter} & 0.669 & 0.712 & 0.771 & \underline{0.869} & 0.568 & 0.550 & 0.717 & 0.792 & 0.699 & 0.746\\
\textbf{FrugalGPT} & 0.657 & \underline{0.719} & 0.562 & 0.622 & 0.551 & 0.552 & 0.504 & 0.515 & 0.545 & 0.575 \\
\textbf{RouteLLM} & 0.641 & 0.681 & 0.492 & 0.485 & 0.480 & 0.475 & 0.499 & 0.498 & 0.532 & 0.513\\
\textbf{MA-GNN} & \underline{0.670} & 0.717 & 0.673 & 0.775 & 0.679 & 0.739 & 0.636 & 0.648 & 0.702 & 0.758 \\
\textbf{TIGER} & 0.571 & 0.567 & 0.739 & 0.735 & 0.656 & 0.691 & 0.639 & 0.683 & 0.710 & 0.764 \\
\hline
\textbf{Ours} & \textbf{0.778} & \textbf{0.843} & \underline{0.774} & \textbf{0.875} & \textbf{0.784} & \textbf{0.859} & \textbf{0.773} & \textbf{0.859} & \textbf{0.771} & \textbf{0.870} \\
\textbf{Ours (new user)} & 0.658 & 0.707 & \textbf{0.780} & 0.858 & \underline{0.759} & \underline{0.824} & \underline{0.756} & \underline{0.833} & \underline{0.751} & \underline{0.831} \\
\Xhline{1pt}
\end{tabular}%
}
\end{table*}

\subsection{Baselines}
We compare our GMTRouter against the following baselines:

\textbf{Prompt-based:} \textbf{(1) Vanilla LLM.} We incorporate the query and the descriptions of candidate LLMs into the prompt, and feed it into LLaMA-3.1-70B \citep{grattafiori2024llama} to select the LLM.
\textbf{(2) Personalized LLM.} Building on the vanilla LLM, we retrieve the ten most relevant interaction histories from the user's training set using semantic query similarity and incorporate them into the prompt. Leveraging in-context learning \cite{dong2022survey}, the LLM is then guided to perform personalized routing.

\textbf{Representative Router: 
(3) GraphRouter. \citep{feng2024graphrouter}} A graph-based model that formulates routing as a node classification task over a graph of queries, tasks, and LLMs with learned edge interactions.
\textbf{(4) FrugalGPT \citep{frugalgpt}} utilizes a PLM to predict the score of the generation result of all LLMs given a query, and then selects the LLM with the highest score within a given cost.
\textbf{(5) RouteLLM \citep{routellm2024}.} Learns to route queries among a weak-strong pair of LLMs. Following the official setup, we designate the weak model as the one with the lower average win rate in the dataset, and the strong model as the one with the higher win rate.

\textbf{Memory-based:}
\textbf{(6) MA-GNN \citep{ma-gnn}.} A memory-augmented GNN that models both \emph{short-term} and \emph{long-term} user interests through item-level message passing and a dedicated memory module. 
\textbf{(7) TIGER \citep{tiger}.} A generative retrieval–based sequential recommender that models item sequences via semantic discrete codes. 

\subsection{Metrics} 
We evaluate the performance of all methods using two metrics: \textbf{(1) Accuracy} measures how often the model correctly identifies the most preferred LLM to answer a given query from a specific user.
\textbf{(2) AUC-ROC} (Area Under the Receiver Operating Characteristic Curve) measures the model's ability to correctly rank candidate LLMs according to user preferences. 

\subsection{Implementation Details} 

 We employ Contriever \citep{izacard2021contriever} as the PLM to obtain the initial node embeddings. We adopt the HGT \citep{hgt} as our model backbone due to its strong capability to maintain dedicated representations for different types of nodes. Additionally, we experiment with various other GNNs as the backbone to investigate their impact on GMTRouter’s performance. Experimental details are provided in Appendix \ref{Sec:gnn-backbones}. We set the visible data size per user to $k=10$ during both training and inference and adopt Entropy Loss as our loss function. In Appendix~\ref{Sec:Visible Data Size}, we will experimentally analyze the impact of different values of $k$ on our method, and hyperparameter details are provided in Appendix \ref{parameter}.
\section{Experiment Results}
\label{Sec:5}


\subsection{Comparison with Baselines}  


We compare GMTRouter with multiple baselines across five datasets in Table \ref{tab:overview}. We observe that GMTRouter outperforms the strongest baseline on \textbf{all evaluation metrics}, achieving up to \textbf{0.108 absolute improvement in accuracy and 0.124 in AUC}, which demonstrates the effectiveness of our framework.
For prompt-based baselines, although incorporating user interaction histories into prompts can improve performance over the Vanilla LLM, these methods still fall significantly behind GMTRouter. This result highlights the limited ability of LLMs to directly extract user preference patterns from raw interaction data.
Moreover, our approach consistently outperforms router-based baselines, including GraphRouter, which has shown strong performance in non-personalized LLM routing tasks. These results validate the importance of leveraging structured information from user--LLM interaction data, together with explicit user preference signals, to better align model selection with diverse user needs. Compared to memory-based methods, GMTRouter benefits from learning over a fine-grained and highly structured heterogeneous graph, leading to superior performance.
Furthermore, even when \textbf{30\% of users are absent from the training set}, GMTRouter remains competitive across datasets, demonstrating its ability to generalize to previously unseen users.

We find that the improvement on ChatBot Arena is relatively modest compared with HUMAINE and the synthetic datasets. ChatBot Arena provides valuable in-the-wild interactions, but its open crowdsourcing setting offers limited control over participant recruitment and interaction quality. Consequently, some comparisons may contain noisy or low-information preference signals, making stable personalized patterns harder to capture. In contrast, HUMAINE recruits compensated participants and applies real-time quality monitoring. Together with its broader candidate-model coverage and more controlled data collection protocol, this provides a more controlled setting for personalized routing, where GMTRouter exhibits a much clearer advantage. We provide further discussion of the selected real-world datasets in Appendix \ref{Sec: Dataset Discussion}.



\begin{figure*}[t]
    \begin{minipage}{\linewidth}
    \vspace{-10pt}
        \centering
        \includegraphics[width=0.45\linewidth]{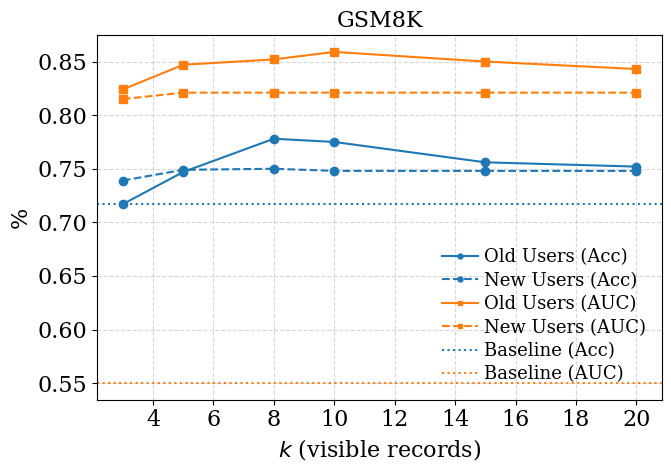}\hfill
        \includegraphics[width=0.45\linewidth]{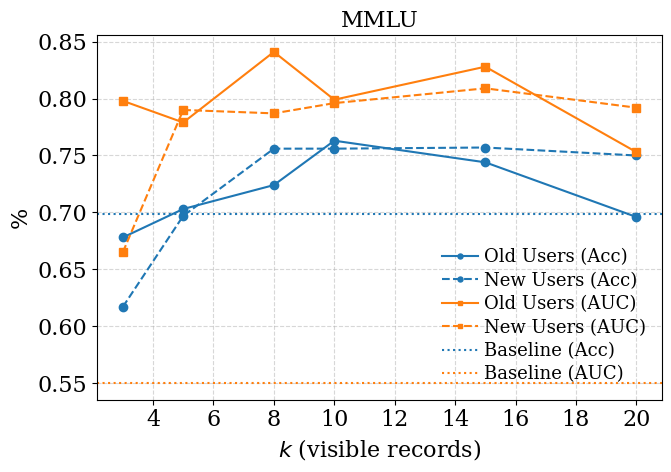}
        
        \captionsetup{type=figure, hypcap=false}
        \phantomsection
        \vspace{-8pt}
        \captionof{figure}{\textbf{Result comparison between old-user and new-user settings for GSM8K (left) and MMLU (right).} The dashed line represents the GraphRouter baseline. The personalized performance under the new-user setting is comparable to that under the old-user setting, highlighting the strong generalization capability of our method.}
        \label{fig:generalization}
    \end{minipage}
\end{figure*}

\begin{table*}[t]
    \centering
    \caption{\textbf{Ablation of design components.}
    We compare the full model with three variants:
    (1) removing the user preference features,
    (2) replacing the prediction head with a dot-product,
    (3) not using user embeddings during prediction.
    The best and second-best results are highlighted in \textbf{bold} and \underline{underline}, respectively.}
    \vspace{-10pt}
    \label{tab:ablation}
    \resizebox{0.9\textwidth}{!}{%
    \begin{tabular}{c||cc|cc|cc|cc}
    \Xhline{1pt}
    \rowcolor{paleaqua}
    \textbf{Method} & \multicolumn{2}{c|}{\textbf{ChatBot Arena}} & \multicolumn{2}{c|}{\textbf{MT-Bench}} & \multicolumn{2}{c|}{\textbf{GSM8K}} & \multicolumn{2}{c}{\textbf{MMLU}} \\
    \rowcolor{paleaqua}
     & Accuracy & AUC & Accuracy & AUC & Accuracy & AUC & Accuracy & AUC \\
    \hline\hline
    w/o $h_p$   & 0.768 & 0.872 & 0.569 & 0.507 & 0.715 & 0.784 & 0.494 & 0.613 \\
    Dot-product   & \textbf{0.777} & 0.868 & \underline{0.730} & \underline{0.795} & 0.629 & 0.724 & 0.681 & 0.746 \\
    w/o $h_u$     & 0.771 & \underline{0.873} & 0.569 & 0.631 & \underline{0.725} & \underline{0.814} & \underline{0.701} & \underline{0.771} \\
    \textbf{GMTRouter} & \underline{0.774} & \textbf{0.875} & \textbf{0.784} & \textbf{0.859} & \textbf{0.773} & \textbf{0.859} & \textbf{0.771} & \textbf{0.870} \\
    \Xhline{1pt}
    \end{tabular}%
    }
\end{table*}

\paragraph{Our Framework is Lightweight}
With only 27.4M trainable parameters and a 109.6MB model size, our framework remains compact compared to existing routing models. During training, only 4.3GB of GPU memory is needed, making it feasible to train on a single modern GPU without specialized hardware. Additional efficiency comparisons are provided in Appendix~\ref{Sec:efficiency-analysis}.






\subsection{Generalization to New Users} \label{Sec:5.3}

We further investigate the personalized capability of our method in few-shot scenarios with new users. Specifically, we evaluate on the GSM8K and MMLU by sampling 30\% users from each dataset and varying the number of visible data $k$. Figure~\ref{fig:generalization} presents averaged results of the sampled users under two settings: (i) the old user setting, where their records are included in the training set, and (ii) the new user setting, where they appear only in the validation and test sets. We observe that new users achieve results comparable to old users, and their performance curves consistently peak far above the GraphRouter baseline. These findings demonstrate that \textbf{our approach effectively learns to capture user preferences from few-shot data and can adapt to new users without requiring extensive fine-tuning.}


\subsection{Ablation Studies}
To evaluate the effectiveness of each design component of the GMTRouter, we conduct ablation studies along the following aspects.

\begin{itemize}[leftmargin=*, nosep]
    \item \textbf{w/o User Preference Feature}
        To verify the effectiveness of the user preference feature in propagating preference signals during GNN aggregation, we remove this feature in this variant. As a result, node embeddings are updated without incorporating preference ratings.
        
    \item \textbf{Dot-product Prediction Head}
        To evaluate whether the cross-attention prediction head captures non-linear interactions more effectively than standard similarity scoring when predicting the optimal model, we replace it in this variant with a simple dot product between the (user + query) and LLM embeddings.

        
    \item \textbf{w/o User Embedding} To evaluate the effectiveness of user embeddings aggregated from the sampled visible graph for personalized prediction, we replace the user embeddings fed into the prediction head with zero vectors, thereby ablating their influence on the predictions.
\end{itemize}

The results of our ablation studies are presented in Table \ref{tab:ablation}. As shown, our GMTRouter achieves the best performance on most metrics compared to the other variants, confirming the effectiveness of our design choices. Additional ablations on multi-turn modeling and turn nodes are provided in Appendix~\ref{Sec:turn-node-ablation}.

\subsection{Further Experiments}

We present additional experiments in Appendix~\ref{Sec: Further Experiment} to further examine the robustness, design choices, generality, and efficiency of GMTRouter. Specifically, we evaluate GMTRouter under more human-like preference factors (Appendix~\ref{Sec: Human-like Preference}), analyze the effects of multi-turn modeling and turn nodes (Appendix~\ref{Sec:turn-node-ablation}), compare against the memory-based MemoryBank baseline (Appendix~\ref{app:memorybank}), and report its computational efficiency (Appendix~\ref{Sec:efficiency-analysis}). We further study the impact of different GNN backbones (Appendix~\ref{Sec:gnn-backbones}) and PLM-based node initialization (Appendix~\ref{Sec: PLM Ablation}).


\section{Additional Related Works}
\vspace{-5pt}

\paragraph{LLM Routing.}
LLM routing focuses on enhancing inference efficiency and response quality by assigning queries to the most appropriate model \citep{masrouter2025,mixllm2025}. Recent work frames routing as learning with cost–quality tradeoffs \citep{kadavath2022know,cascadeRouting2024}: RouteLLM learns from preference data ~\cite{routellm2024}, and RouterBench offers standardized routing benchmarks ~\cite{routerbench2024}. BEST-Route jointly selects LLM and generation count at test-time via a bandit controller ~\cite{bestroute2025}. However, existing approaches are not fully personalized and fail to exploit user information from interaction histories as well as the structure of multi-turn dialogues.

\paragraph{Heterogeneous Graph Learning.}
HetGNNs are designed to model heterogeneous graphs by capturing complex multi-type interactions among various nodes and edges \citep{allset2021,hgnn2019}. HAN uses hierarchical attention over metapaths ~\cite{han2019}, while MAGNN and HeCo improve metapath aggregation and cross-view contrast ~\cite{magnn2020,heco2021}. Transformers such as HGT provide inductive, relation-aware message passing with temporal encoding ~\cite{hgt}. This enables rich relational structures in user–LLM interactions while leveraging inductive training to enhance generalization on sparse data from new users. 




\vspace{-5pt}
\section{Conclusion}
 In this work, we introduced GMTRouter, a heterogeneous graph-based framework for personalized LLM routing. By modeling multi-turn user–LLM interactions as a heterogeneous graph and propagating preference signals across node types, our method effectively captures user-specific patterns even from few-shot data. Experiments across five benchmarks confirm that GMTRouter consistently surpasses strong baselines, while adapting efficiently to new users without retraining. These results highlight the value of structured interaction modeling for advancing preference-aware LLM routing and point to promising future directions in scalable, user-aligned LLM deployment.

\newpage
\section{Limitations}
While GMTRouter demonstrates significant improvements in personalized routing, we identify a few limitations that offer avenues for future research. First, the initial embeddings for LLM nodes are currently derived from static descriptions provided by API platforms. While effective, these representations may require manual updates if a model undergoes significant versioning or performance shifts. Second, although we evaluated our framework across five diverse benchmarks spanning reasoning, math, and general knowledge, its performance in highly specialized niche domains, such as advanced medical or legal sub-specialties, remains to be further explored. Finally, the current implementation focuses on individual user personalization; extending the framework to model collective or group-level preferences in collaborative environments represents a promising direction for future work.

\section{Ethical Considerations}

Our experiments rely on publicly released datasets and do not involve new human-subject data collection. HUMAINE includes demographic attributes, which we use only to construct aggregate preference cohorts; we do not attempt to identify individual participants. Personalized routing may nevertheless inherit or amplify biases present in user preference data and candidate LLMs, and demographic cohorting may obscure preference heterogeneity within each group. In practical deployment, user interaction histories may also contain sensitive information. Appropriate consent, data minimization, access control, and privacy protections should therefore be considered when applying personalized routing systems to real users.

\bibliography{custom}

\appendix

\clearpage
\onecolumn

\section*{Appendix Contents}

\begingroup
\small
\setlength{\parskip}{2pt}

\newcommand{\appsec}[2]{%
    \noindent\textbf{\ref{#1}\quad #2}\dotfill\pageref{#1}\par
}

\newcommand{\appsubsec}[2]{%
    \noindent\hspace{1.5em}\ref{#1}\quad #2\dotfill\pageref{#1}\par
}

\appsec{Implementation Details}{Implementation Details}
\appsubsec{parameter}{Model Configuration and Hyperparameters}
\appsubsec{app:sampling-unit}{Sampling Unit in User-conditioned Graph Sampling}
\appsubsec{app:training}{Training of GMTRouter}

\vspace{4pt}
\appsec{Dataset Preparation}{Dataset Preparation}
\appsubsec{Sec: Dataset Details}{Details for Datasets}
\appsubsec{sec: dataset}{Dataset Statistics}
\appsubsec{app:arena-scale}{Scale of the ChatBot Arena Motivation Analysis}
\appsubsec{app:feedback-modalities}{Feedback Modalities and Sparse Supervision}
\appsubsec{sec:user weight}{Synthetic User Design}
\appsubsec{Sec: Dataset Discussion}{Discussion of Real-World Datasets}

\vspace{4pt}
\appsec{Sec:Visible Data Size}{Investigating the Impact of Visible Data Size}

\vspace{4pt}
\appsec{app:new-user-results}{Additional Results on New User Experiments}

\vspace{4pt}
\appsec{Sec: Further Experiment}{Further Experiments}
\appsubsec{Sec: Human-like Preference}{Human-like Preference Augmentation}
\appsubsec{Sec:turn-node-ablation}{Effect of Multi-turn Modeling and Turn Nodes}
\appsubsec{app:memorybank}{Comparison with MemoryBank}
\appsubsec{Sec:efficiency-analysis}{Efficiency Analysis}
\appsubsec{Sec:gnn-backbones}{Using Different GNNs as the GMTRouter Backbone}
\appsubsec{Sec: PLM Ablation}{Ablation on PLM-based Node Initialization}

\vspace{4pt}
\appsec{app:baseline-prompts}{Baseline Routing Prompts}
\appsubsec{app:personalized-llm-retrieval}{Retrieval Details for the Personalized LLM Baseline}
\appsubsec{app:routing-prompt-templates}{Routing Prompt Templates}

\vspace{4pt}
\appsec{app:llm-use}{The Use of Large Language Models (LLMs)}

\endgroup

\clearpage
\twocolumn

\section{Implementation Details}
\label{Implementation Details}

\subsection{Model Configuration and Hyperparameters} \label{parameter}

We implement our method using PyTorch Geometric \citep{Fey/Lenssen/2019} and conduct all experiments on a single NVIDIA RTX A6000 GPU.

\paragraph{Architecture.}
We use a heterogeneous graph transformer (HGT) with:
\begin{itemize}[leftmargin=*, nosep]
  \item \textbf{GNN}: 2 layers (single-turn) or 3 (multi-turn), 768-dim hidden, 4-head HGTConv, LayerNorm, dropout 0.1.
  \item \textbf{Predictor}: 4-head cross-attention followed by an MLP with hidden dimension 256 and dropout 0.1, where the LLM representation attends to the user and query representations.
\end{itemize}

\paragraph{Training.}
\begin{itemize}[leftmargin=*, nosep]
  \item \textbf{Epochs}: 1000 \quad \textbf{LR}: 5e-4
  \item \textbf{Visible records/user} ($k$): \{3, 5, 8, 10, 15, 20\}
  \item \textbf{Batch size}: 256 supervision triplets
  \item \textbf{Model selection}: prioritize validation AUC, then Accuracy.
\end{itemize}

\subsection{Sampling Unit in User-conditioned Graph Sampling}
\label{app:sampling-unit}

In user-conditioned graph sampling, the basic sampling unit is one user--query instance from the Interaction History Table. Each instance contains a user, a query or multi-turn query context, a set of candidate LLMs, the corresponding responses generated by these candidate LLMs, and the user feedback associated with these responses. Therefore, sampling $k$ interaction histories for a user means selecting $k$ such user--query instances from this user's available interaction records to construct the visible subgraph for message passing.

This design matches the personalized routing objective: for a given user and query, the router estimates the user's preference over candidate LLMs and selects the LLM with the highest predicted preference score. In our experiments, each sampled instance contains two candidate LLM responses. The visible subgraph constructed from these sampled instances provides the local evidence used to infer the user's preference representation during both training and inference.

\subsection{Training of \textbf{GMTRouter}}
\label{app:training}

Algorithm~\ref{alg:training} summarizes the training procedure of GMTRouter.

\begin{algorithm}[h]
\small
\caption{Training \textbf{GMTRouter}}
\label{alg:training}
\DontPrintSemicolon
\KwData{$\mathcal{D}_{\mathrm{train}}=\{(x,y)\}$}
\KwParams{epochs $E$, visible $k$, supervision $s$, PLM, GNN $f_\phi$, predictor $\mathrm{Pred}$}
\KwInit{PLM-encode query, response, and LLM nodes; initialize user/turn nodes and edge features}

\For{$e \gets 1$ \KwTo $E$}{
  $\mathcal{G}^{(e)} \gets$ subgraph from $k|\mathcal{U}|$ visible records\;
  $\mathcal{M}^{(e)} \gets s$ held-out triples $(u,q,m)$\;
  $h \gets f_\phi(\mathcal{G}^{(e)})$ \tcp*[r]{message passing}
  \For{$(u,q,m)\in\mathcal{M}^{(e)}$}{
    $\hat y \gets \mathrm{Pred}(h_u,\, h_q,\, h_m)$\;
  }
  Update $f_\phi$ and $\mathrm{Pred}$ by minimizing $\mathcal{L}_{\mathrm{rank}}(\hat y,y)$\;
}
\end{algorithm}

\section{Dataset Preparation}
\label{Dataset Preparation}

\subsection{Details for datasets}\label{Sec: Dataset Details}
We evaluate our approach on both real-world and synthetic datasets spanning five distinct tasks to provide a comprehensive assessment.
\begin{itemize}[leftmargin=*, nosep]
    \item \textbf{HUMAINE \citep{petrova2026unpackinghumanpreferencellms}:} HUMAINE is a large-scale benchmark of naturalistic multi-turn interactions with human preference judgments across demographically stratified participants. We group records sharing the same demographic profile into a preference cohort and treat each cohort as one ``user''. For our experiments, we select the 20 ``users'' with the most interaction records. Detailed statistics are provided in Appendix \ref{sec: dataset}.
    \item \textbf{ChatBot Arena \citep{Chiang2024Chatbot}:} ChatBot Arena is a real-world dataset of anonymized multi-turn conversations with pairwise human preference labels across LLMs. For our experiments, we select the 11 users and 16 LLMs with the largest numbers of interactions. Detailed statistics are provided in Appendix \ref{sec: dataset}.
    \item \textbf{MT-Bench \citep{zheng2023judging}:} MT-Bench is a benchmark for evaluating the reasoning and multi-turn conversational capabilities of LLMs, containing 80 multi-turn questions.
    \item \textbf{GSM8K \citep{cobbe2021gsm8k}:} GSM8K is a dataset of grade school-level math word problems, designed to assess LLMs’ mathematical reasoning and problem-solving skills.
    \item \textbf{MMLU \citep{hendrycks2021ethics,hendryckstest2021}:} MMLU is a comprehensive benchmark covering 57 subjects across STEM, humanities, social sciences, and professional domains, used to measure general knowledge and multi-domain reasoning abilities of LLMs. We sample 10 questions from each subject for our experiments.
\end{itemize}

\subsection{Dataset Statistics} \label{sec: dataset}

We preprocess each dataset into user--query--LLM--response records and partition them into training, validation, and test sets. Table~\ref{tab:dataset_stats} summarizes the resulting statistics. We evaluate generalization to unseen users separately under the new-user setting described in Section~\ref{Sec:5.3}.

\begin{table}[h]
\centering
\caption{Dataset statistics, including the number of entries, users, and LLMs in each split.}
\label{tab:dataset_stats}
\begin{tabular}{llccc}
\toprule
\textbf{Dataset} & \textbf{Split} & \textbf{Entries} & \textbf{Users} & \textbf{LLMs} \\
\midrule
ChatBot  & Train & 2780 & 11 & 16 \\
Arena  & Valid & 386  & 11 & 16 \\
  & Test  & 824  & 11 & 16 \\
\midrule
 MT & Train & 2240 & 10 & 2 \\
 Bench & Valid & 320  & 10 & 2 \\
  & Test  & 640  & 10 & 2 \\
\midrule
\multirow{3}{*}{GSM8K} 
  & Train & 18460 & 10 & 2 \\
  & Valid & 2620 & 10 & 2 \\
  & Test  & 5300 & 10 & 2 \\
\midrule
\multirow{3}{*}{MMLU} 
  & Train & 3970 & 5 & 2 \\
  & Valid & 560  & 5 & 2 \\
  & Test  & 1150  & 5 & 2 \\
\midrule
\multirow{3}{*}{HUMAINE} 
  & Train & 3535 & 20 & 45 \\
  & Valid & 483  & 20 & 45 \\
  & Test  & 1020 & 20 & 45 \\
\bottomrule
\end{tabular}
\end{table}

For ChatBot Arena, we selected the following users and LLMs:

\textbf{Users:} arena\_user\_9965, arena\_user\_15085, arena\_user\_257, arena\_user\_13046, arena\_user\_11473, arena\_user\_3820, arena\_user\_9676, arena\_user\_6467, arena\_user\_6585, arena\_user\_5203, arena\_user\_1338

\textbf{LLMs:} koala-13b, vicuna-13b, gpt-3.5-turbo, oasst-pythia-12b, gpt-4, claude-v1, RWKV-4-Raven-14B, palm-2, alpaca-13b, mpt-7b-chat, vicuna-7b, claude-instant-v1, chatglm-6b, fastchat-t5-3b, dolly-v2-12b, stablelm-tuned-alpha-7b

\begin{table*}[t]
\centering
\caption{Computational cost of graph construction on four datasets.}
\resizebox{\textwidth}{!}{%
\begin{tabular}{lcccc}
\toprule
\textbf{Dataset} & \textbf{Data Entries} & \textbf{Avg.\ Tokens} & \textbf{Encoding Time} (s) & \textbf{Graph Construction Time} (s) \\
\midrule
ChatBot Arena & 3990  & 184.41  & 51.73 & 1.70 \\
MT-Bench      & 3200  & 4511.73 & 55.68 & 2.40 \\
GSM8K         & 26380 & 112.68  & 142.84 & 1.49 \\
MMLU          & 5680  & 9.35    & 4.27 & 1.56 \\
\bottomrule
\end{tabular}
}
\label{tab:graph_cost}
\end{table*}

We report in Table \ref{tab:graph_cost} the size of each dataset along with the time required to process it into the heterogeneous graph used in our experiments.

\begin{table}[h]
\centering
\small
\caption{Number of interaction records for the 10 active ChatBot Arena users used in the motivation analysis.}
\label{tab:arena_motivation_scale}
\resizebox{0.7\columnwidth}{!}{%
\begin{tabular}{lc}
\toprule
\textbf{User ID} & \textbf{\# Records} \\
\midrule
arena\_user\_110 & 51 \\
arena\_user\_15085 & 288 \\
arena\_user\_11941 & 84 \\
arena\_user\_14189 & 79 \\
arena\_user\_257 & 292 \\
arena\_user\_973 & 99 \\
arena\_user\_4715 & 69 \\
arena\_user\_9965 & 376 \\
arena\_user\_12353 & 64 \\
arena\_user\_13514 & 78 \\
\midrule
\textbf{Total} & \textbf{1,480} \\
\bottomrule
\end{tabular}
}
\end{table}

\subsection{Scale of the ChatBot Arena Motivation Analysis}
\label{app:arena-scale}

In Section~\ref{sec:2.2}, we use a subset of 10 active ChatBot Arena users to analyze cross-user differences in LLM preferences. This subset is used only for the motivation analysis and is distinct from the ChatBot Arena split used in the main experiments. Although each selected user has at least 50 records, the analysis covers 1,480 interaction records in total, as shown in Table~\ref{tab:arena_motivation_scale}.

\subsection{Feedback Modalities and Sparse Supervision}
\label{app:feedback-modalities}

GMTRouter is designed to support heterogeneous feedback formats rather than assuming that all interactions share the same type of supervision. In our formulation, user feedback may appear as scalar ratings, pairwise or listwise preference rankings, or ground-truth responses. These feedback modalities can be converted into comparable preference signals over candidate LLM responses: scalar ratings are used directly, rankings are discretized into relative preference scores, and ground-truth responses are converted into supervision by measuring the similarity between the generated response and the reference response.

This design allows GMTRouter to be applied to datasets with different forms of personalization supervision. For example, HUMAINE and ChatBot Arena provide pairwise human preferences, whereas the synthetic datasets use user-specific ratings constructed from multiple response-level attributes, as described in Appendix~\ref{sec:user weight}.

We also note that GMTRouter does not require dense feedback histories for each user. During both training and inference, user-conditioned graph sampling constructs a visible subgraph from only a small number of sampled interaction histories, as detailed in Appendix~\ref{app:sampling-unit}. This enables the model to infer user preferences from sparse supervision and supports personalization in few-shot or cold-start settings.

\subsection{Synthetic User Design} \label{sec:user weight}

To simulate diverse user preferences, we introduce synthetic users whose routing behavior is governed by a weighted linear utility function over multiple response-level attributes: response quality, generation cost, response length, and rare-word count. These attributes are designed to reflect individualized trade-offs, such as prioritizing answer quality, preferring lower-cost responses, favoring concise versus detailed answers, or preferring simple versus more technical language. For each dataset, we manually assign different weights $\{w_{\text{quality}}, w_{\text{cost}}, w_{\text{length}}, w_{\text{rare}}\}$ per user to simulate these personalized preferences. These weights are normalized within each dataset to prevent scale bias.

\begin{table}[h]
\centering
\caption{\textbf{Synthetic user weights for MT-Bench dataset.}}
\label{tab:synthetic_users_mt}
\resizebox{\columnwidth}{!}{
\begin{tabular}{lrrrr}
\toprule
\textbf{User} & $w_{\text{quality}}$ & $w_{\text{length}}$ & $w_{\text{rare}}$ & $w_{\text{cost}}$ \\
\midrule
user\_1  & 1.42 &  0.0087  & $-0.174$ & $-45.23$ \\
user\_2  & 1.87 &  0.0012  &  0.091   & $-15.55$ \\
user\_3  & 0.96 &  0.0135  &  0.045   & $-48.42$ \\
user\_4  & 1.15 & $-0.0008$ & $-0.220$ & $-10.00$ \\
user\_5  & 1.69 &  0.0024  &  0.175   & $-38.50$ \\
user\_6  & 1.08 & $-0.0015$ & $-0.030$ & $-25.12$ \\
user\_7  & 0.53 &  0.0162  &  0.230   & $-5.75$  \\
user\_8  & 1.34 & $-0.0005$ & $-0.145$ & $-12.40$ \\
user\_9  & 1.98 &  0.0101  &  0.087   & $-25.10$ \\
user\_10 & 1.57 &  0.0024  & $-0.065$ & $-7.79$  \\
\bottomrule
\end{tabular}
}
\end{table}

\begin{table}[h]
\centering
\caption{\textbf{Synthetic user weights for GSM8K dataset.}}
\label{tab:synthetic_users_gsm8k}
\begin{tabular}{lrrrr}
\toprule
\textbf{User} & $w_{\text{quality}}$ & $w_{\text{length}}$ & $w_{\text{rare}}$ & $w_{\text{cost}}$ \\
\midrule
user\_1  & 1.0 & 20.0 & 100.0 & -0.0 \\
user\_2  & 1.5 & 18.0 & 50.0 & -1.0 \\
user\_3  & 0.8 & 22.0 & 80.0 & -0.5 \\
user\_4  & 1.2 & 17.0 & 120.0 & -0.2 \\
user\_5  & 2.0 & 15.0 & 70.0 & -0.4 \\
user\_6  & 0.4 & 6.0 & -4.0 & -1.0 \\
user\_7  & 0.3 & 7.0 & -5.0 & -0.9 \\
user\_8  & 0.6 & 8.0 & -7.0 & -1.2 \\
user\_9  & 0.2 & 9.0 & -9.0 & -0.8 \\
user\_10 & 0.8 & 10.0 & -3.0 & -1.1 \\
\bottomrule
\end{tabular}
\end{table}

\begin{table}[h]
\centering
\caption{\textbf{Synthetic user weights for MMLU dataset.}}
\label{tab:synthetic_users_mmlu}
\begin{tabular}{lrrrr}
\toprule
\textbf{User} & $w_{\text{quality}}$ & $w_{\text{length}}$ & $w_{\text{rare}}$ & $w_{\text{cost}}$ \\
\midrule
user\_1 & 1.0 & 0.00 & 0.00 &     0.0 \\
user\_2 & 1.0 & 0.00 & 0.00 & $-600.0$ \\
user\_3 & 1.0 & 0.00 & 0.00 & $-1200.0$ \\
user\_4 & 1.0 & 0.00 & 0.00 & $-1800.0$ \\
user\_5 & 1.0 & 0.00 & 0.00 & $-2400.0$ \\
\bottomrule
\end{tabular}
\end{table}






\begin{figure*}[t]
    \begin{minipage}{\linewidth}
        \centering
        \includegraphics[width=0.45\linewidth]{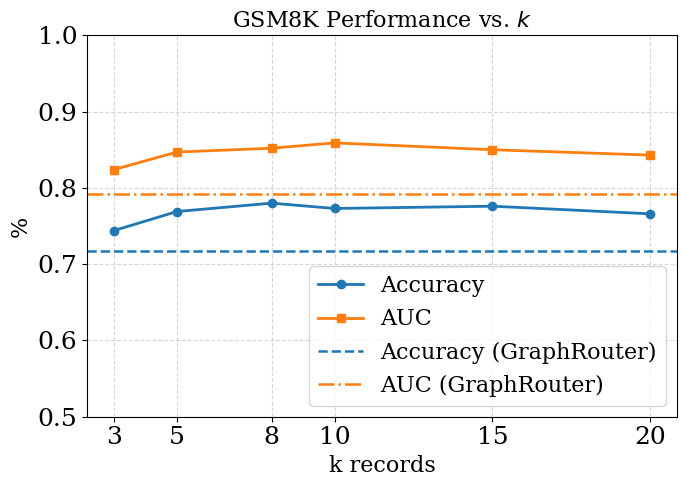}\hfill
        \includegraphics[width=0.45\linewidth]{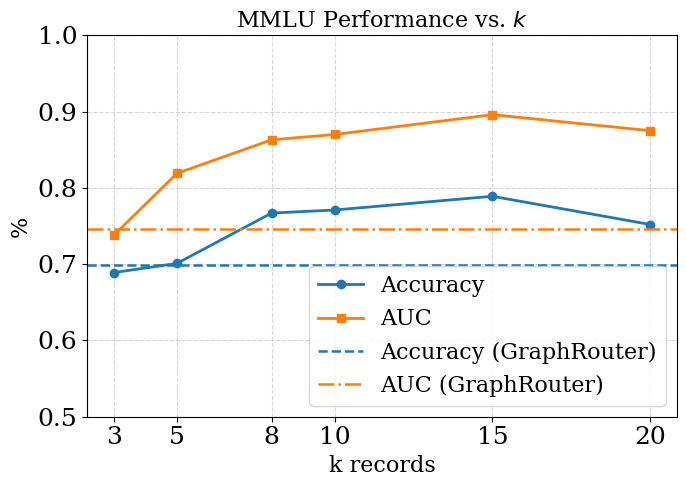}
        \caption{This figure illustrates \textbf{the impact of the visible data size $k$ on GMTRouter for GSM8K (left) and MMLU (right).} The dashed line represents the GraphRouter baseline. As $k$ increases, the performance of our method improves, but it saturates once $k$ reaches 10.}
        \label{fig:k-selection}
    \end{minipage}
\end{figure*}

\begin{figure*}[t]
    \begin{minipage}{\linewidth}
        \centering
        \includegraphics[width=0.49\linewidth]{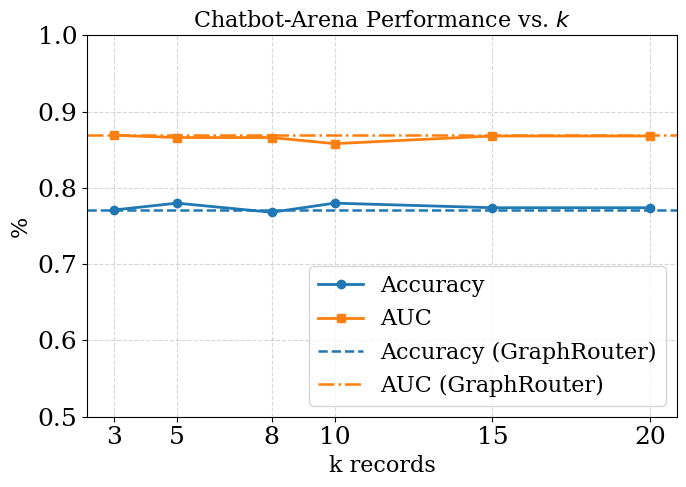}\hfill
        \includegraphics[width=0.49\linewidth]{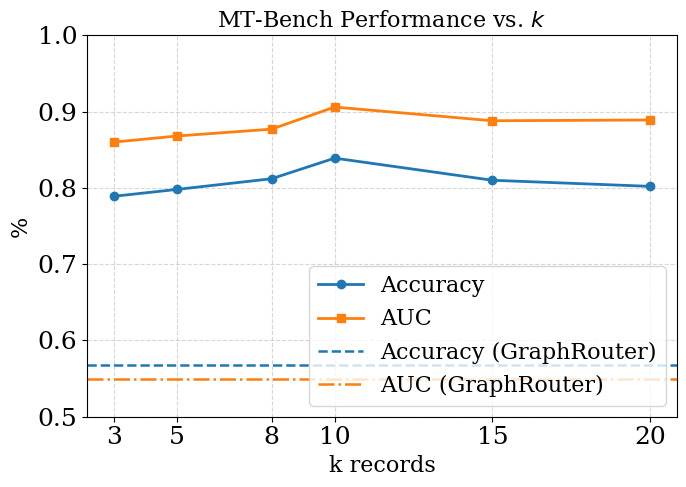}
        \caption{This figure illustrates \textbf{the impact of the visible data size $k$ on GMTRouter for ChatBot Arena (left) and MT-Bench (right).} The dashed line represents the GraphRouter baseline. As $k$ increases, the performance of our method improves, but it saturates once $k$ reaches 10.}
        \label{fig:k-selection_appendix}
    \end{minipage}
\end{figure*}


\subsection{Discussion of Real-World Datasets}
\label{Sec: Dataset Discussion}

Both HUMAINE and ChatBot Arena contain real-world multi-turn interactions with direct human preference judgments, but they differ substantially in their data collection protocols and degree of quality control.

\paragraph{ChatBot Arena.}
ChatBot Arena \citep{Chiang2024Chatbot} is collected through open online crowdsourcing, where anonymous, self-selected users freely submit prompts and vote between model responses. This setting provides valuable in-the-wild interactions, but offers limited control over participant recruitment and interaction quality. Consequently, some comparisons can contain noisy or low-information preference signals, making stable personalized patterns harder to identify. Our manual inspection finds examples where the preferred response is objectively incorrect or where both responses substantially fail the task. Representative cases are shown in Table~\ref{tab:chatbot_noise}.

\begin{table*}[t]
\centering
\small
\begin{tabular}{p{0.20\linewidth} p{0.22\linewidth} p{0.22\linewidth} p{0.28\linewidth}}
\toprule
\textbf{Query} & \textbf{Preferred response} & \textbf{Rejected response} & \textbf{Analysis} \\
\midrule
``Write a function in Scheme that reverses a list of strings?'' &
\texttt{dolly-v2-12b}: \texttt{(define reverse-list ... (list (car l) (reverse-list (cdr l))))} &
\texttt{gpt-3.5-turbo}: \texttt{(define (reverse-list-of-strings lst) ... (append ...))} &
The preferred response is incomplete and constructs nested lists, whereas the rejected response at least provides executable recursive code that reverses the list order. \\
\midrule
``Simulate Python console. Input: \texttt{print([2 * 2] * 2)}'' &
\texttt{dolly-v2-12b}: \texttt{[4, 4, 4, 4]} &
\texttt{fastchat-t5-3b}: \texttt{print(2 * 2 * 2)} &
The correct output is \texttt{[4, 4]}; both responses are incorrect, making the preference signal uninformative. \\
\midrule
``For each skill, choose their favorite horror game: Logic, Encyclopedia, Rhetoric, Drama, ...'' &
\texttt{chatglm-6b}: \texttt{Logic: ``The Shining'', ... Encyclopedia: ``The Great Gatsby'', ...} &
\texttt{mpt-7b-chat}: \texttt{Logic: Inland Empire; Encyclopedia: Shivers; ...} &
The preferred response mainly lists movies, books, or unrelated titles rather than horror games, while the rejected response maps skills to other skill names. Both substantially fail the task. \\
\bottomrule
\end{tabular}
\caption{Representative low-information or potentially unreliable preference comparisons in ChatBot Arena.}
\label{tab:chatbot_noise}
\end{table*}

We emphasize that these examples do not imply that ChatBot Arena is generally low-quality. Rather, they illustrate an inherent trade-off of its open, in-the-wild collection setting: broad coverage and natural user behavior come with noisier preference supervision. This may partially explain why GMTRouter shows a more modest improvement on ChatBot Arena than on the other datasets.

\paragraph{HUMAINE.}
HUMAINE \citep{petrova2026unpackinghumanpreferencellms} provides a more controlled real-user evaluation setting. Participants are recruited and compensated through Prolific, each comparison requires at least three conversational turns, and low-effort or disengaged interactions are monitored in real time. The benchmark also contains substantially more interaction records and candidate LLMs than the ChatBot Arena subset used in our experiments. These properties provide cleaner and larger-scale preference supervision for personalized routing, under which GMTRouter achieves substantially larger improvements over strong baselines.

HUMAINE does not release persistent participant identifiers. Therefore, as described in Section~\ref{Sec: Dataset Details}, we group records sharing the same demographic profile into a preference cohort and treat each cohort as one ``user''. This construction captures demographic-level preference heterogeneity rather than longitudinal preferences of a single individual, and should be interpreted accordingly.

\section{Investigating the Impact of Visible Data Size}
\label{Sec:Visible Data Size}
We investigate the impact of $k$ visible data per user on the quality of the aggregated node embeddings. Figures \ref{fig:k-selection} and \ref{fig:k-selection_appendix} present the test results on four datasets, respectively. As $k$ increases, both accuracy and AUC improve, but beyond $k{=}10$, the performance begins to plateau or slightly decline, indicating diminishing returns from including additional visible data. This may be due to reduced generalization or potential instability caused by excessively large batch sizes during training \citep{Keskar2016On,Oyedotun2022A}. Therefore, we choose $k{=}10$ as a balanced setting for capturing user preferences without compromising generalization.

\section{Additional Results on New User Experiments}
\label{app:new-user-results}

\begin{figure*}
    \begin{minipage}{\linewidth}
        \centering
        \includegraphics[width=0.49\linewidth]{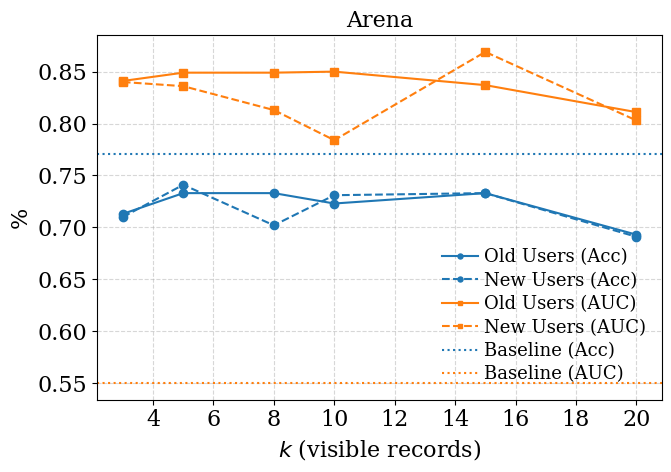}\hfill
        \includegraphics[width=0.49\linewidth]{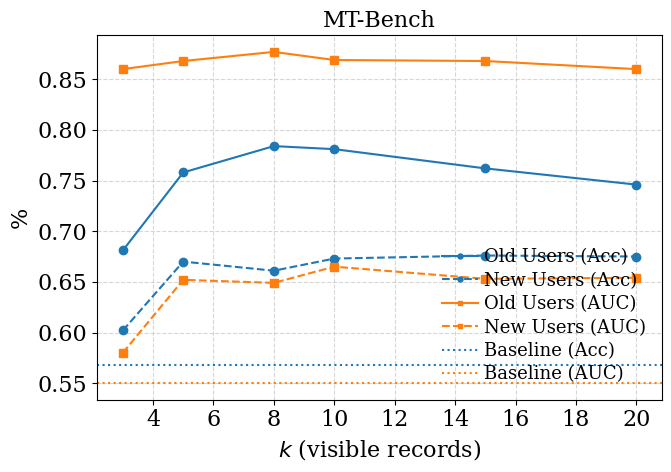}
        
        \captionsetup{type=figure, hypcap=false}
        \phantomsection
        \captionof{figure}{\textbf{Result comparison between old-user and new-user settings for ChatBot Arena (left) and MT-Bench (right).} The dashed line represents the GraphRouter baseline. The personalized performance under the new-user setting is comparable to that under the old-user setting, highlighting the strong generalization capability of our method.}
        \label{fig:generalization_appendix}
    \end{minipage}
\end{figure*}

Here, we present the results of the experiment described in Section \ref{Sec:5.3} on the ChatBot Arena and MT-Bench datasets, as shown in Figure \ref{fig:generalization_appendix}.

\section{Further Experiments}\label{Sec: Further Experiment}

\subsection{Human-like Preference Augmentation}
\label{Sec: Human-like Preference}

To further examine whether GMTRouter can capture richer user preferences beyond the original proxy metrics, we construct an augmented synthetic personalization setting on MT-Bench with additional human-like preference factors. In the original synthetic setting, user preferences are simulated using response quality, cost, response length, and rare-word count. While these dimensions capture important aspects of user preferences, they may not fully reflect more subjective factors such as tone, safety, and factuality.

Specifically, we use Qwen3-32B-AWQ as an annotator to label each response with three additional attributes: (1) \textbf{tone}, selected from a predefined set of response styles; (2) \textbf{safety}, represented as a binary label; and (3) \textbf{factuality}, represented as a score from 0 to 100. We then extend each simulated user profile with four additional preference attributes: a preferred tone, a tone-matching weight, a safety weight, and a factuality weight. These additional attributes are combined with the original synthetic preference dimensions to reconstruct the user-specific ratings.

Table~\ref{tab:human_like_preference} reports the results on this augmented MT-Bench setting. GMTRouter consistently outperforms all representative baselines, achieving the best performance in both Accuracy and ROC-AUC. These results suggest that GMTRouter remains effective when user preferences involve more human-like and subjective factors beyond the original proxy metrics.

\begin{table}[t]
    \centering
    \caption{\textbf{Performance on the augmented MT-Bench setting with additional human-like preference factors.}
    The best results are highlighted in \textbf{bold}.}
    \vspace{-5pt}
    \label{tab:human_like_preference}
    \resizebox{0.8\columnwidth}{!}{%
    \begin{tabular}{l||cc}
    \Xhline{1pt}
    \rowcolor{paleaqua}
    \textbf{Method} 
    & \textbf{Accuracy} 
    & \textbf{AUC} \\
    \hline\hline
    GraphRouter 
    & 0.667 & 0.717 \\
    TIGER 
    & 0.642 & 0.721 \\
    RouteLLM 
    & 0.429 & 0.462 \\
    FrugalGPT 
    & 0.642 & 0.687 \\
    GMTRouter 
    & \textbf{0.715} & \textbf{0.777} \\
    \Xhline{1pt}
    \end{tabular}%
    }
\end{table}

\subsection{Effect of Multi-turn Modeling and Turn Nodes}
\label{Sec:turn-node-ablation}

To evaluate the effect of multi-turn modeling and the proposed turn-node design, we conduct additional ablation experiments on MT-Bench. We compare three settings: (1) \textbf{Single-turn only}, where the model is trained and evaluated using only single-turn interactions; (2) \textbf{Multi-turn w/o turn nodes}, where the model uses the same multi-turn data but removes turn nodes from the heterogeneous graph; and (3) \textbf{Multi-turn}, which corresponds to the full GMTRouter setting with explicit turn nodes.

Table~\ref{tab:turn-node-ablation} summarizes the results. Multi-turn interactions substantially improve routing performance over the single-turn setting, indicating that richer interaction context provides useful preference signals for personalized routing. Moreover, adding turn nodes further improves performance over the multi-turn variant without turn nodes, especially in ROC-AUC. This supports our design motivation that turn nodes help aggregate local information within each dialogue round and facilitate preference propagation across multi-turn user--LLM interactions.

\begin{table}[h]
    \centering
    \caption{\textbf{Effect of multi-turn modeling and turn nodes on MT-Bench.}
    The best and second-best results are highlighted in \textbf{bold} and \underline{underline}, respectively.}
    \vspace{-5pt}
    \label{tab:turn-node-ablation}
    \resizebox{\columnwidth}{!}{%
    \begin{tabular}{l||cc}
    \Xhline{1pt}
    \rowcolor{paleaqua}
    \textbf{Setting} & \textbf{Accuracy} & \textbf{AUC} \\
    \hline\hline
    Single-turn only & 0.463 & \underline{0.709} \\
    Multi-turn w/o turn nodes & \underline{0.644} & 0.669 \\
    Multi-turn & \textbf{0.675} & \textbf{0.786} \\
    \Xhline{1pt}
    \end{tabular}%
    }
\end{table}

\subsection{Comparison with MemoryBank}
\label{app:memorybank}

To further compare GMTRouter with recent memory-based personalization methods, we include MemoryBank~\citep{zhong2024memorybank} as an additional baseline. MemoryBank equips LLMs with a long-term memory module that stores, retrieves, and updates past interactions to support personalized responses over time. We adapt it to the personalized routing setting by retrieving user-specific interaction histories as memory and using the backbone LLM to select the candidate LLM that best matches the user's preference. We use Llama-4-Scout-17B-16E-Instruct as the backbone model for this baseline.

Table~\ref{tab:memorybank_results} reports the results across four datasets. GMTRouter consistently outperforms MemoryBank on both Accuracy and ROC-AUC. This comparison further supports our main observation that, compared with methods that primarily store and retrieve unstructured user histories, GMTRouter benefits from learning over a fine-grained heterogeneous graph that explicitly models user--query--LLM--response interactions.

\begin{table*}[t]
    \centering
    \caption{\textbf{Comparison with MemoryBank across four datasets.}
    The best results are highlighted in \textbf{bold}.}
    \vspace{-5pt}
    \label{tab:memorybank_results}
    \resizebox{\textwidth}{!}{%
    \begin{tabular}{l||cc|cc|cc|cc}
    \Xhline{1pt}
    \rowcolor{paleaqua}
    \textbf{Method} 
    & \multicolumn{2}{c|}{\textbf{ChatBot Arena}} 
    & \multicolumn{2}{c|}{\textbf{MT-Bench}} 
    & \multicolumn{2}{c|}{\textbf{GSM8K}} 
    & \multicolumn{2}{c|}{\textbf{MMLU}}  \\
    \rowcolor{paleaqua}
    \textbf{Metric} 
    & Accuracy & AUC 
    & Accuracy & AUC 
    & Accuracy & AUC 
    & Accuracy & AUC  \\
    \hline\hline
    MemoryBank~\citep{zhong2024memorybank} 
    & 0.601 & 0.634 
    & 0.575 & 0.555 
    & 0.555 & 0.552 
    & 0.610 & 0.604  \\
    GMTRouter 
    & \textbf{0.774} & \textbf{0.875} 
    & \textbf{0.784} & \textbf{0.859} 
    & \textbf{0.773} & \textbf{0.859} 
    & \textbf{0.771} & \textbf{0.870}  \\
    \Xhline{1pt}
    \end{tabular}%
    }
\end{table*}

\subsection{Efficiency Analysis}
\label{Sec:efficiency-analysis}

To further evaluate the computational overhead of GMTRouter, we compare it with representative training-based routing baselines on MT-Bench. We report training time, peak training GPU memory, and inference latency per query in Table~\ref{tab:efficiency_analysis}. Since LLM-based prompting baselines require substantially larger inference-time computation, we focus this comparison on lightweight training-based methods.

The results show that GMTRouter introduces additional training cost due to its graph-structured user--LLM interaction modeling. However, its peak training memory remains close to other lightweight training-based routers, and its inference latency is still at the millisecond level. This indicates that GMTRouter preserves practical efficiency while providing stronger personalized routing performance.

\begin{table*}[h]
    \centering
    \caption{\textbf{Efficiency comparison on MT-Bench.}
    Lower values indicate better efficiency, and the best results are highlighted in \textbf{bold}.}
    \vspace{-5pt}
    \label{tab:efficiency_analysis}
    \resizebox{0.8\textwidth}{!}{%
    \begin{tabular}{l||ccc}
    \Xhline{1pt}
    \rowcolor{paleaqua}
    \textbf{Method} & \textbf{Training Time} & \textbf{Peak Train GPU Mem} & \textbf{Inference Time / Query} \\
    \rowcolor{paleaqua}
    & \textbf{(s)} & \textbf{(MB)} & \textbf{(s)} \\
    \hline\hline
    GraphRouter & 156.77 & 1898.64 & 0.0052 \\
    TIGER & \textbf{32.65} & \textbf{1475.35} & \textbf{0.0014} \\
    GMTRouter & 710.46 & 2050.70 & 0.0063 \\
    \Xhline{1pt}
    \end{tabular}%
    }
\end{table*}

\subsection{Using Different GNNs as the GMTRouter Backbone}
\label{Sec:gnn-backbones}

We investigate how different GNNs used as the GMTRouter backbone affect its personalization capability. We evaluate four backbone models, including both homogeneous and heterogeneous GNNs: GraphSAGE \citep{Hamilton2017Inductive}, HGT \citep{hgt}, HAN \citep{han2019}, and HeteroConv, and present the results in Table \ref{tab:gnn-backbones}.

HGT achieves the strongest overall performance, obtaining the best AUC across all four datasets and the best or near-best Accuracy. HAN is also competitive on several datasets. These results further indicate that GMTRouter is not dependent on any particular GNN architecture: multiple attention-based backbones achieve strong performance, suggesting that the gains primarily stem from the graph-based personalized routing framework and data modeling rather than from a specific convolutional operator.

\begin{table*}[t]
    \centering
    \caption{\textbf{GMTRouter performance with different GNN backbones.}
    The best and second-best results are highlighted in \textbf{bold} and \underline{underline}, respectively.}
    \vspace{-5pt}
    \label{tab:gnn-backbones}
    \resizebox{\textwidth}{!}{%
    \begin{tabular}{c||cc|cc|cc|cc}
    \Xhline{1pt}
    \rowcolor{paleaqua}
    \textbf{Backbone} & \multicolumn{2}{c|}{\textbf{ChatBot Arena}} & \multicolumn{2}{c|}{\textbf{MT-Bench}} & \multicolumn{2}{c|}{\textbf{GSM8K}} & \multicolumn{2}{c}{\textbf{MMLU}} \\
    \rowcolor{paleaqua}
    \textbf{Metric} & Accuracy & AUC & Accuracy & AUC & Accuracy & AUC & Accuracy & AUC \\
    \hline\hline
    GraphSAGE     & 0.768 & \underline{0.873} & 0.569 & 0.645 & 0.635 & 0.648 & 0.494 & 0.487 \\
    HeteroConv    & \textbf{0.777} & 0.867 & 0.569 & 0.492 & 0.499 & 0.603 & 0.494 & 0.542 \\
    HANConv       & 0.766 & 0.776 & \underline{0.646} & \underline{0.680} & \textbf{0.774} & \underline{0.775} & \underline{0.707} & \underline{0.746} \\
    HGTConv       & \underline{0.774} & \textbf{0.875} & \textbf{0.784} & \textbf{0.859} & \underline{0.773} & \textbf{0.859} & \textbf{0.771} & \textbf{0.870} \\
    \Xhline{1pt}
    \end{tabular}%
    }
\end{table*}

\subsection{Ablation on PLM-based Node Initialization}
\label{Sec: PLM Ablation}

GMTRouter uses PLM embeddings to initialize the query, response, and LLM nodes, thereby incorporating their semantic information into the interaction graph. To examine the contribution of such semantic initialization, we construct a variant that replaces these PLM embeddings with non-semantic 64-dimensional hash embeddings while keeping the remaining architecture unchanged.

As shown in Table~\ref{tab:hash_embedding_results}, replacing PLM representations with hash embeddings consistently degrades performance across all four datasets, reducing Accuracy by \textbf{0.058--0.129} and AUC by \textbf{0.077--0.146}. This confirms that semantic node initialization provides useful information for personalized routing. Nevertheless, PLM initialization serves as a supporting component of GMTRouter; the core design of our framework remains the structured modeling of multi-turn user--LLM interactions together with user-conditioned inductive graph learning.

\begin{table*}[t]
    \centering
    \caption{\textbf{Ablation on PLM-based node initialization across four datasets.}
    We replace the PLM embeddings of query, response, and LLM nodes with non-semantic 64-dimensional hash embeddings. The best results are highlighted in \textbf{bold}.}
    \vspace{-5pt}
    \label{tab:hash_embedding_results}
    \resizebox{\textwidth}{!}{%
    \begin{tabular}{l||cc|cc|cc|cc}
    \Xhline{1pt}
    \rowcolor{paleaqua}
    \textbf{Method}
    & \multicolumn{2}{c|}{\textbf{HUMAINE}}
    & \multicolumn{2}{c|}{\textbf{MT-Bench}}
    & \multicolumn{2}{c|}{\textbf{GSM8K}}
    & \multicolumn{2}{c}{\textbf{MMLU}} \\
    \rowcolor{paleaqua}
    \textbf{Metric}
    & Accuracy & AUC
    & Accuracy & AUC
    & Accuracy & AUC
    & Accuracy & AUC \\
    \hline\hline
    Hash Embedding
    & 0.649 & 0.697
    & 0.658 & 0.729
    & 0.702 & 0.782
    & 0.713 & 0.767 \\
    GMTRouter
    & \textbf{0.778} & \textbf{0.843}
    & \textbf{0.784} & \textbf{0.859}
    & \textbf{0.773} & \textbf{0.859}
    & \textbf{0.771} & \textbf{0.870} \\
    \Xhline{1pt}
    \end{tabular}%
    }
\end{table*}

\section{Baseline Routing Prompts}
\label{app:baseline-prompts}

\subsection{Retrieval Details for the Personalized LLM Baseline}
\label{app:personalized-llm-retrieval}

For the Personalized LLM baseline, retrieval is performed separately for each user rather than globally. Given a new query from a user, we first encode the query using Contriever. We then encode all historical queries from the same user's training set and compute their inner-product similarity with the new query embedding. The top-10 most similar interaction histories are retrieved and inserted into the prompt as personalized context. This per-user retrieval design ensures that the prompt contains only interaction histories from the target user, allowing the LLM baseline to perform personalized routing based on user-specific preference evidence.

\subsection{Routing Prompt Templates}
\label{app:routing-prompt-templates}

To benchmark routing strategies, we design two representative prompt templates: one for a vanilla router that selects the best LLM without personalization, and another for a personalized router that incorporates user history and preferences. Both prompts simulate realistic routing scenarios where a system must choose a single LLM for the next turn in a multi-turn dialogue.

\section{The Use of Large Language Models (LLMs)}
\label{app:llm-use}

During the writing of this paper, we used the GPT-5 Mini model for text polishing and grammatical corrections to enhance the readability of the manuscript.

\begin{table*}[t]
\vspace{-35pt}
\centering
\caption{\small \textbf{Prompt Template: Vanilla LLM Routing (No Personalization)}}
\label{prompt-template:vanilla-routing}
\begin{tabular}{p{13cm}}
\toprule[1.1pt]
\textbf{[Instruction]} \\

You are an expert routing agent. Your task is to select the most suitable Large Language Model (LLM) to handle the next query in a multi-turn conversation. \\

\textbf{[Input Format]} \\
\texttt{[Candidate LLM List]} \\
\texttt{\{\{CANDIDATE\_LLM\_LIST\}\}} \\

\texttt{[Previous Conversation]} \\
\texttt{\{\{PREVIOUS\_CONVERSATION\}\}} \\

\texttt{[Current Query]} \\
\texttt{\{\{CURRENT\_QUERY\}\}} \\

\textbf{[Instructions for Model Selection]} 
\begin{itemize}
  \item Consider the query difficulty, the context of the previous conversation, and each LLM’s expertise, cost, and size.
  \item Choose the single best LLM to respond to the current query.
  \item Output only the name of the selected LLM in the exact format below.
  \item Do not provide explanations or commentary.
\end{itemize}

\textbf{[Output Format]} \\
\texttt{<'\{selected\_model\_name\}'>}
\\
\bottomrule[1.1pt]
\end{tabular}
\end{table*}


\begin{table*}[h]
\centering
\caption{\small \textbf{Prompt Template: Personalized Routing (User History Aware)}}
\label{prompt-template:personalized-routing}
\begin{tabular}{p{13cm}}
\toprule[1.1pt]
\textbf{[Instruction]} \\

You are an expert routing agent. Your task is to select the most suitable Large Language Model (LLM) to handle the next query in a multi-turn conversation, incorporating both model characteristics and personalization signals from the user’s history. \\

\textbf{[Input Format]} \\
\texttt{[Candidate LLM List]} \\
\texttt{\{\{CANDIDATE\_LLM\_LIST\}\}} \\

\texttt{[Previous Conversation]} \\
\texttt{\{\{PREVIOUS\_CONVERSATION\}\}} \\

\texttt{[Current Query]} \\
\texttt{\{\{CURRENT\_QUERY\}\}} \\

\texttt{[User Preference History]} \\
\texttt{\{\{USER\_PREFERENCE\_HISTORY\}\}} \\

\textbf{[Instructions for Model Selection]} 
\begin{itemize}
  \item Consider the query difficulty, the context of the ongoing conversation, the LLMs’ specializations, cost, and size.
  \item Additionally, factor in the user’s historical preferences and ratings to personalize the routing decision.
  \item Choose the single best LLM to respond to the current query.
  \item Output only the name of the selected LLM in the exact format below.
  \item Do not provide explanations or commentary.
\end{itemize}

\textbf{[Output Format]} \\
\texttt{<'\{selected\_model\_name\}'>}
\\
\bottomrule[1.1pt]
\end{tabular}
\end{table*}

\end{document}